\documentclass{article}
\usepackage{amsmath,epsfig}
\usepackage[preprint]{spconfa4}
\usepackage{subfigure}
\usepackage{multirow}
\usepackage{booktabs}
\usepackage{makecell}
\usepackage{cite}
\usepackage{color}
\usepackage{algorithmic}
\usepackage{graphicx}
\usepackage{textcomp}
\usepackage{xcolor}
\usepackage{placeins}
\usepackage{float}
\usepackage{tabularx,colortbl}
\usepackage{hyperref}
\hypersetup{
    colorlinks=true,
    linkcolor=blue,
    filecolor=magenta,      
    urlcolor=cyan,
}

\let\OLDthebibliography\thebibliography
\renewcommand\thebibliography[1]{
	\OLDthebibliography{#1}
	\setlength{\parskip}{0pt}
	\setlength{\itemsep}{0pt plus 0.3ex}
}

\begin{document}\sloppy

	% Example definitions.
	% --------------------
	\def\x{{\mathbf x}}
	\def\L{{\cal L}}

	% Title.
	% ------
	\title{Edge-directed Geometric Partitioning for Versatile Video Coding}
	%
	% Address.
	% ---------------
	\name{Xuewei Meng$^{\ast}$, Xinfeng Zhang$^{\dagger}$, Chuanmin Jia$^{\ast}$, Xia Li$^{\ddagger}$, Shanshe Wang$^{\ast}$, and Siwei Ma$^{\ast}$}
	\address{$^{\ast}$Institute of Digital Media, Peking University, Beijing, China\\
		$^{\dagger}$University of Chinese Academy of Sciences, Beijing, China\\
		$^{\ddagger}$Shenzhen Graduate School, Peking University, Shenzhen, China\\
		Email: \{xwmeng, cmjia, ethanlee, sswang, swma\}@pku.edu.cn, xfzhang@ucas.ac.cn}

	\maketitle

	\begin{abstract}
		To improve the coding performance, geometric partition~(GEO) was proposed for the upcoming VVC standard. GEO provides 140 partition candidates. The index of optimal GEO mode needs to be signaled explicitly. Considering different structural characteristics of different CUs and the correlation between spatial adjacent blocks and temporal collocated	blocks, we propose a GEO mode prediction strategy by constructing a Most Probable Mode~(MPM) list to reduce the overhead of GEO index and improve coding efficiency. Based on the observation of the high correlation between the partition mode and object boundaries, an edge-directed geometric partition scheme is proposed to construct the MPM list according to spatio-temporal edge information. The proposed method provides an objective BD-rate gain of 0.58\% and 1.00\% on average for RA and LDB configurations compared to VTM-6.0. Besides, it also promotes the visual quality of object boundaries.
	\end{abstract}
	\section{Introduction}
	\label{sec:intro}
	The block-based coding structure has been recognized as the core of the state-of-the-art video coding standards, such as H.264/AVC~\cite{AVC} and H.265/HEVC~\cite{HEVC}, because of its capability in achieving high compression efficiency. In AVC, a tree-structured variable block-size partition was employed for motion estimation~(ME) and motion compensation~(MC) to approximate the motion vector~(MV) fields within each $16 \times 16$ macroblock~(MB). To further improve the flexibility of block partition, quadtree-based~(QT) block partition structure~\cite{kim2012block} and non-square QT structure~\cite{NSQT} were introduced in HEVC. Then, the quadtree with a nested multi-type tree~(QT-MTT) using binary tree and ternary tree partition structure~\cite{chen2018description} replaced the conventional QT partition in Versatile Video Coding~(VVC), a new video coding standard, developed by Joint Video Experts Team~(JVET)~\cite{CFP} since October 2017.
	
	Although the partition structure becomes more flexible and efficient, the rectangular partition in the current video coding standard is not able to represent the arbitrary shape of an actual object with a uniform MV in a rectangular coding block. This constraint limits the performance of inter prediction and may cause discontinuity on object edges. To address this issue, many non-rectangular partition methods were developed in the context of AVC and HEVC. To improve the performance of motion compensation, a wedge-based geometric partition~(WBP) approach~\cite{kondo2005motion} was proposed to generate a better partition of the discontinuous MV fields by splitting a $16\times16$ MB into two wedge-shaped parts using a straight line. In which, the encoder performed ME on each part to find the best-matched corresponding reference block, and each of the two resulting wedge-shaped parts was predicted with an associated MV. This kind of non-rectangular partition method could achieve high coding efficiency, but also dramatically increase the encoder complexity since more recursive splitting iterations were required. Although many optimization approaches~\cite{ferreira2009efficiency, muhit2009fast, guo2010simplified} were proposed to reduce the encoder complexity, there was still a long way for this technology to be adopted by video coding standards because of its lousy trade-off between the complexity and coding efficiency.
	
\begin{figure}[t!]
		\begin{center}
			\noindent
			\includegraphics[width=0.36\linewidth]{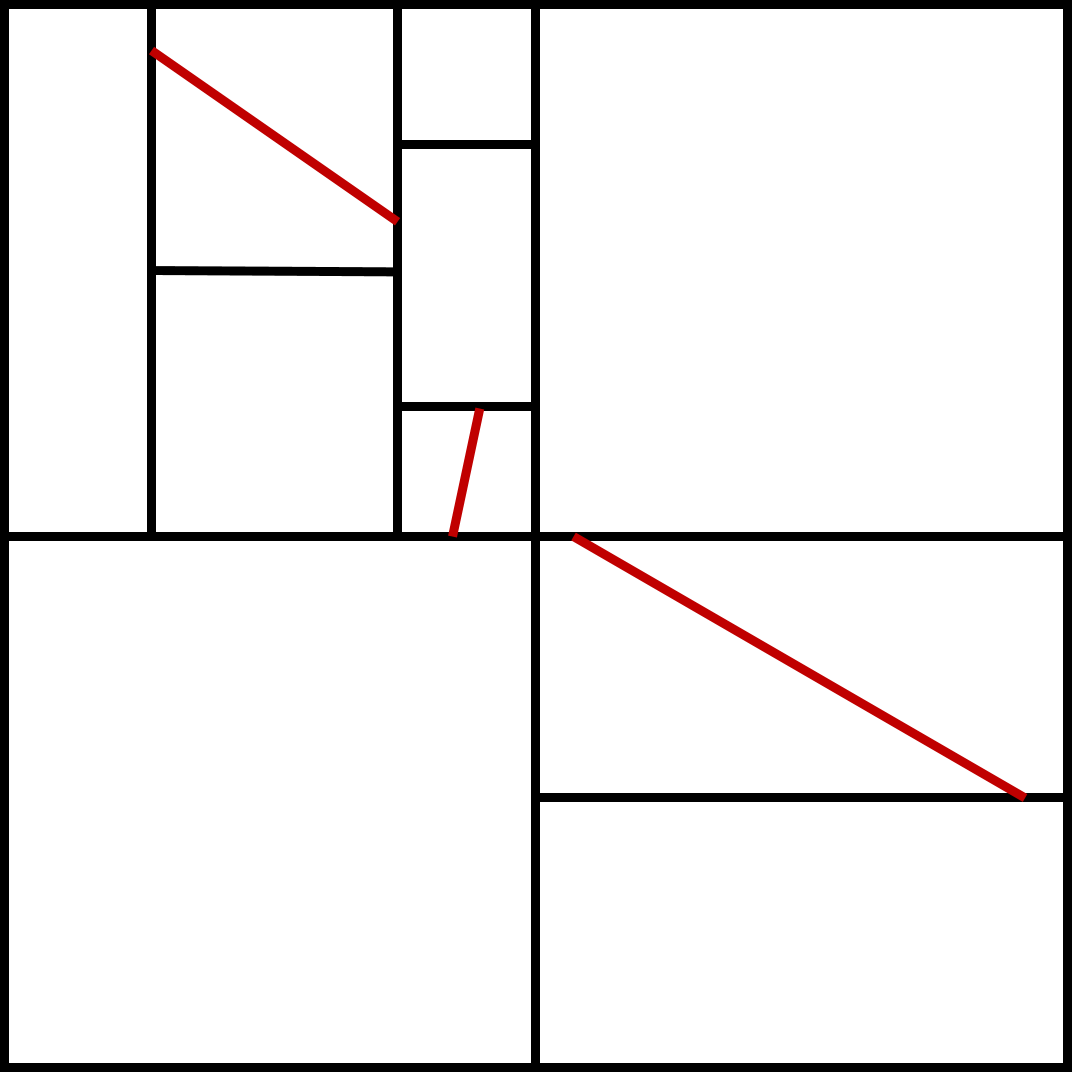}
			\caption{An example of QT-MTT based geometric partition scheme. All leaf node coding units of QT-MTT can be further geometrically partitioned by a straight line.}\label{partitioning}
		\end{center}
		\vspace{-2.5mm}
	\end{figure}

	In VVC, these partition methods have also been investigated. In~\cite{L0125,L0417}, geometric partition based on the QT-MTT partition structure was proposed, which is shown in Fig.~\ref{partitioning}. Although the complexity still remained comparable with that in AVC and HEVC, the coding performance was less efficient because of the more flexible QT-MTT partition structure adopted in VVC. To achieve a better trade-off between the performance and complexity, JVET-L0124~\cite{L0124}, which has been adopted in VVC, proposed only to apply triangle partition~(TPM) for merge mode. By doing so, the most time-consuming process, ME for geometric partition, was removed in the encoder. The adoption of TPM improved the representation capability of inter prediction, but the arbitrary object shape was still challenging to approximate with the pre-defined triangular and rectangular coding unit~(CU). To further improve coding efficiency, a new geometric partition scheme~(GEO)~\cite{P0068} for merge mode with up to 140 partition candidates was proposed. In GEO, one partition mode was selected for a CU from the pre-defined 140 candidates, and an index~($0\sim139$) was signaled to indicate the selected mode. Due to different structural characteristics of different CUs and the correlation between spatial adjacent blocks and temporal collocated blocks, it was unnecessary to use 140 partition modes for each CU. Moreover, the side information of GEO index~(up to 8 bits) was also a significant burden of overhead which existed the potential for optimization. 
	
	In this paper, we propose a GEO mode prediction strategy by constructing a Most Probable Mode~(MPM) list in the encoder and the decoder to reduce the overhead of GEO index and improve coding efficiency. Based on the observed high correlation between the partition mode and object boundaries, an edge-directed geometric partition scheme~(E-GEO) is proposed to construct the MPM list according to spatio-temporal predicted edge information. The main contributions are summarized as follows,
	
	\begin{itemize}
		\item We propose a GEO mode prediction strategy by using an MPM list in encoder and decoder. To the best of our knowledge, it is the first MPM-based geometric partition framework for inter prediction.
		
		\item We propose a spatio-temporal edge detection method to construct the MPM list. Spatial and temporal pixels are jointly used to estimate the edge map of current CU. 
	\end{itemize}
	
	\section{Edge-directed geometric partition}
	In this section, we firstly review the conventional GEO algorithm, then analyze the distribution of geometric partition in the decoder, which expertly guides us to design the proposed framework. Finally, the proposed edge-directed geometric partition method is elaborated.	
	
	\subsection{Geometric partition framework}
	\begin{figure}[ht!]
		\begin{center}
			\noindent
			\includegraphics[width=2.55in]{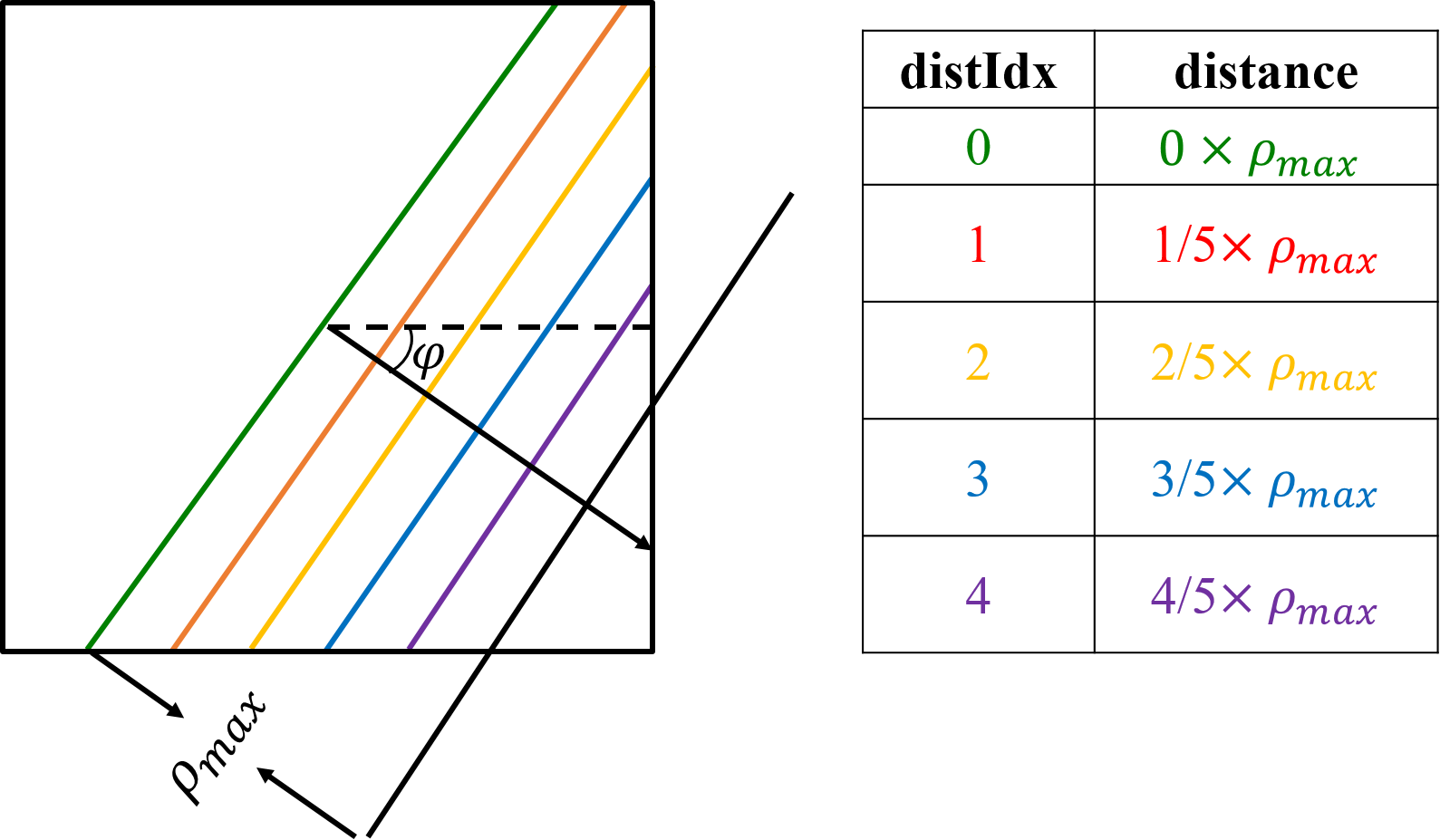}
			\caption{Distance offset of GEO partition mode.}\label{distance}
		\end{center}
		\vspace{-2.5mm}
	\end{figure}

	To achieve a better trade-off between coding efficiency and the computational complexity, the GEO partition scheme proposed in this paper is only applied for merge mode. The first step of GEO is the Merge Candidate List construction like other merge modes, such as regular merge mode and affine merge mode. Secondly, the best partition mode is selected from the GEO mode set for each CU. And the best Motion Vectors corresponding to the two partitions of each CU are determined from Merge Candidate List. Then, motion compensation is conducted by using the best MVs derived in the previous step, and geometric partition edge blending is performed according to the selected partition mode to get the final reconstructed CU. Details will be shown in Section.~\ref{mode_sets} and Section.~\ref{blendingProcess}.
	
	\subsubsection{GEO mode set} \label{mode_sets}
	The split boundary of geometric merge mode is described by the angle $\varphi_i$~($i = 0, 1, ...~(I - 1)$) and the distance offset $\rho_j$~($j = 0, 1, ...,~(J - 1)$). Here, $\varphi_i$ represents a quantized angle between 0 and 360 degrees with step 11.25 degree~($I = 32$). The distance offset $\rho_j$ represents a quantized offset of the largest $\rho_{max}$ with a fixed step~($J = 5$). In addition, split directions overlapped with QT-MTT and TPM splits are excluded. Hence, there are 140 modes in total.
	
	For the case that $\varphi$ is equal to 0, $\rho_{max}$ is equal to $w/2$. If $\varphi$ is equal to 90 degree, $\rho_{max}$ is set to be $h/2$. The “1.0” sample backshift is to avoid that the split boundary is too close to the corner. When $0<\varphi<\frac{\pi}{2} \label{ro}$, $\rho_{max}$ can be formulated as~\cite{P0068}, 
	
	\begin{equation}\label{ro}
	\rho_{max}(\varphi, w, h) = cos(\varphi)\times(\frac{h}{2tan(\frac{\pi}{2} - \varphi)} + \frac{w}{2}) - 1.0,
	\end{equation}
	
	\noindent where $h$ and $w$ are height and width of the current CU, respectively. As shown in Fig.~\ref{distance}, for a given $\varphi$, $\rho_{max}$ is calculated according to Eq.~\eqref{ro}. Then, 5 distances corresponding to $0$, $1/5\times\rho_{max}$, $2/5\times\rho_{max}$, $3/5\times\rho_{max}$, $4/5\times\rho_{max}$ are derived.
	
	\subsubsection{Blending along the geometric partition edge}\label{blendingProcess}
\begin{figure}[ht!]
		\begin{center}
			\noindent
			\includegraphics[width=2.6in]{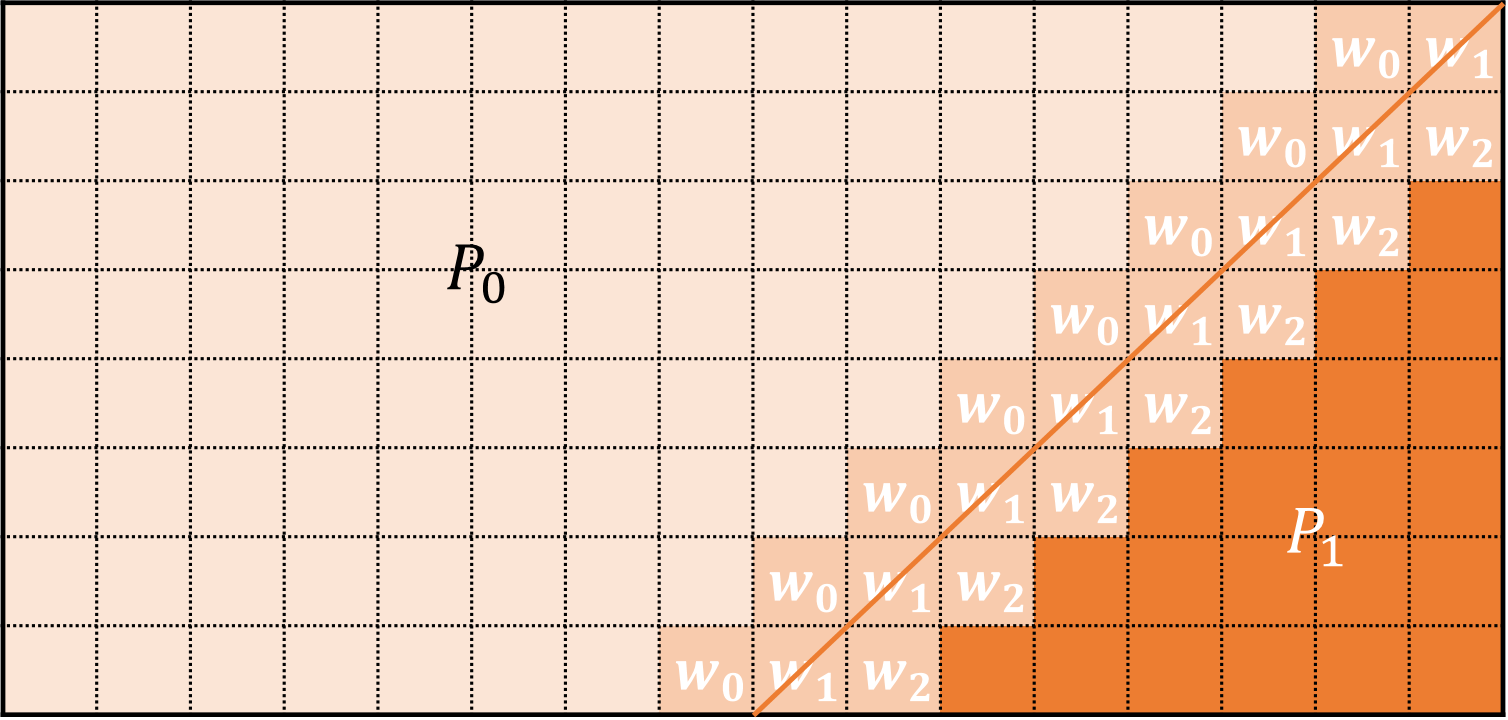}
			\caption{An example of blending process for $16\times8$ CU.}\label{blending}
		\end{center}
		\vspace{-1.0mm}
	\end{figure}

	After predicting each non-rectangular partition using its own motion vector~($MV_0$ for $P_0$ and $MV_1$ for $P_1$), blending is applied to the two prediction blocks to derive samples around the partition edge, which is shown in Fig.~\ref{blending}. The blending process can be represented as follows,
	
	\begin{equation}
	P_B = (w\times{P_0} + (8 - w)\times{P_1} + 4) >> 3 \label{blend_weight},
	\end{equation}
	
	\noindent where $P_0$ and $P_1$ are blocks reconstructed by using $MV_0$ and $MV_1$, respectively. $w$ corresponding to $w_0$, $w_1$ and $w_2$ in Fig.~\ref{blending} is the blending weight, derived by using look-up tables~\cite{P0068}. For the non-blending area, the samples in $P_0$ or $P_1$ are used as the final reconstructed samples directly. For the blending area, reconstructed samples are derived by Eq.~\eqref{blend_weight}.
	
	\makeatletter
	\newcommand{\thickhline}{%
		\noalign {\ifnum 0=`}\fi \hrule height 1pt
		\futurelet \reserved@a \@xhline
	}
	
	\subsection{GEO mode analysis}
\begin{figure}[!hb]
		\begin{center}
			\noindent
			\subfigure[\textit{RacehorsesC}~(QP=37, the $58^{th}$ frame, RA)]{
				\includegraphics[width=2.6in]{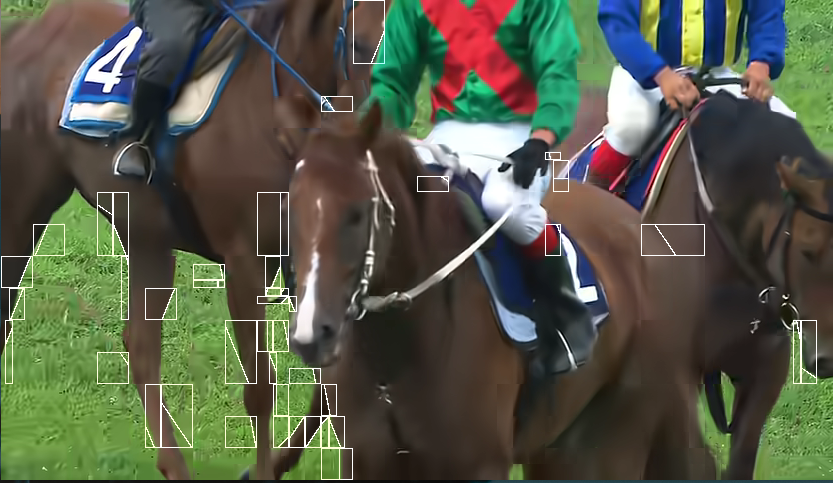}
			}
			\subfigure[\textit{BQMall}~(QP=32, the $110^{th}$ frame, LDB)]{
				\includegraphics[width=2.6in]{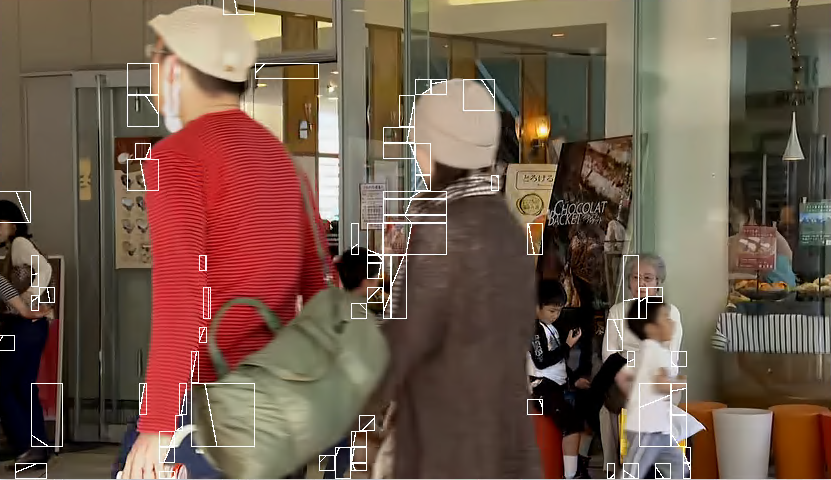}
			}
			\caption{Examples of GEO.}\label{dec_partition}
		\end{center}
		\vspace{-1.0mm}
	\end{figure}

	To analyze the influencing factors of GEO mode decision, we estimate the mode selection frequency in decoder, as shown in Fig.~\ref{dec_partition}. GEO is selected mostly on the object edge, and the partition mode has a strong correlation with object boundaries. As we mentioned before, the large quantity of overhead caused by the GEO partition index is the primary reason for the limited coding efficiency. So, we propose to guide the MPM list construction using object edge information.
	
	\subsection{Edge-directed geometric partition framework}
\begin{figure}[t!]
		\begin{center}
			\noindent
			\includegraphics[width=3.35in]{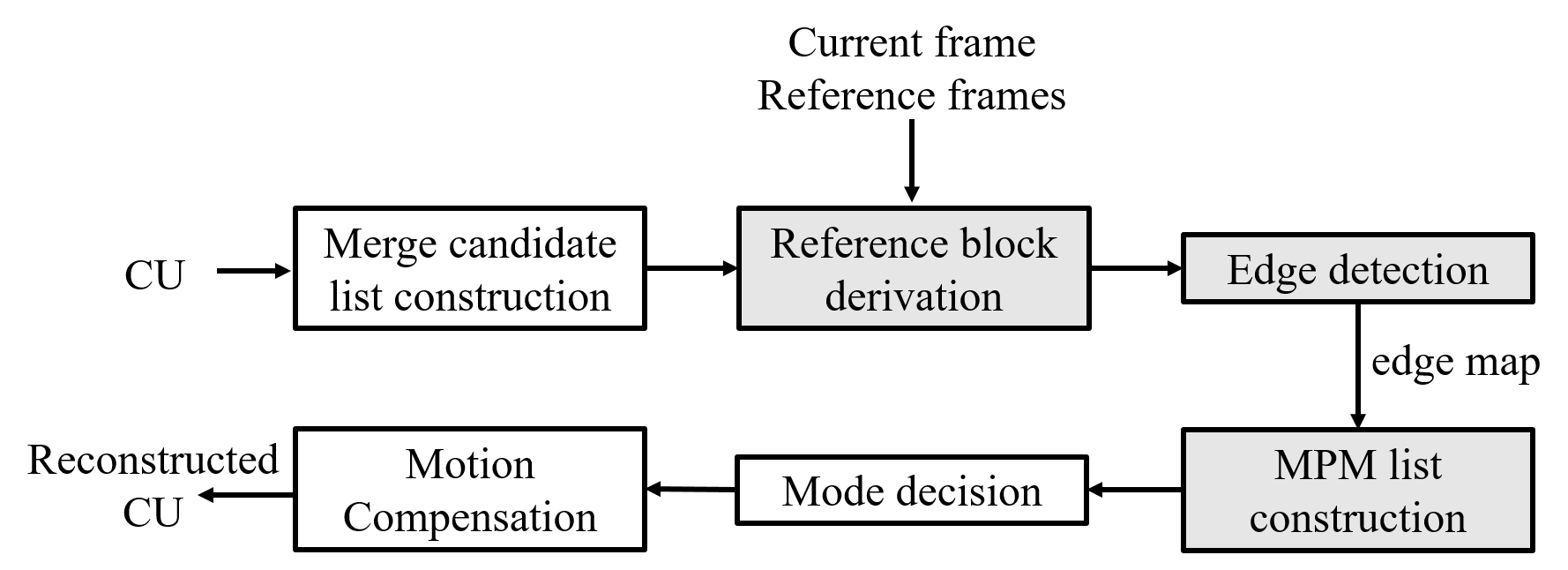}
			\caption{The framework of the proposed E-GEO scheme in encoder, with the gray boxes highlighting our contributions.}\label{pro-frame}
		\end{center}
		\vspace{-1.0mm}
	\end{figure}

	To derive the MPM of current CU, edge estimation is needed in the proposed method. Moreover, as the original frame is unavailable at decoder, we can only use the reconstructed area of the current frame and reference frames to estimate the edge of current CU. As shown in Fig.~\ref{pro-frame}, spatio-temporal based reference block derivation process is conducted firstly. Then guided filter~\cite{he2012guided} and canny detector~\cite{canny1986computational} are used for edge detection. Finally, the MPM list is constructed according to the predicted edge map.
	
	\subsubsection{Reference block derivation}
\begin{figure}[h]
		\begin{center}
			\noindent
			\includegraphics[width=3.35in]{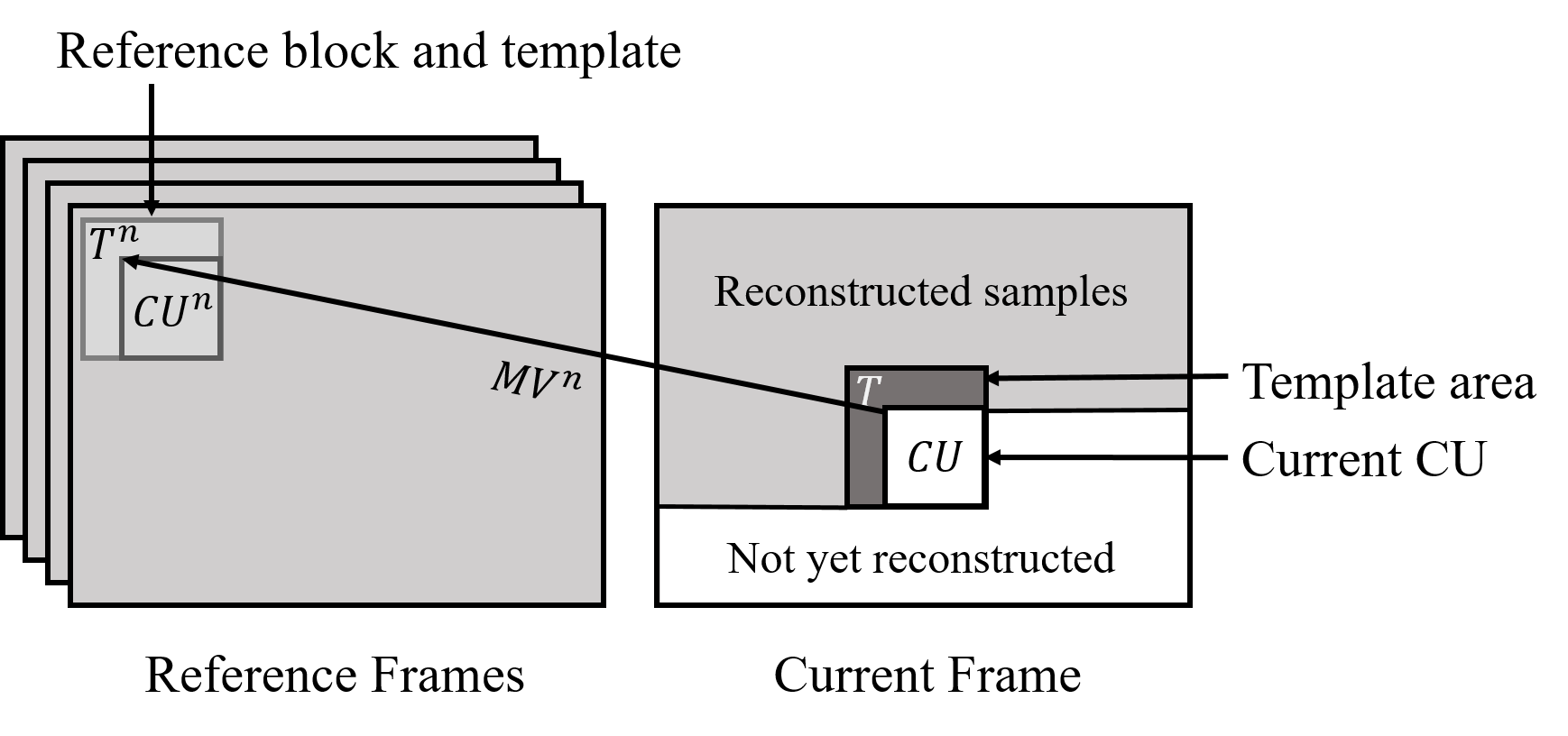}
			\caption{Reference block derivation process.}\label{templateMatching}
		\end{center}
		\vspace{-1.0mm}
	\end{figure}

	The reference block is derived according to spatio-temporal information. First, we derive the $n$-th MV~($MV_n$) from GEO merge candidate list, and round it to integer-pixel accuracy to avoid the interpolation process. Then, MSD~(Mean of Squared Difference) between the template area of current CU~($T$) and that of its reference block~($T^{n}$) is used as the similarity evaluation criteria, as shown in Fig.~\ref{templateMatching}. $CU$ and $T$ represent the current CU and its template area. $CU^{n}$ and $T^{n}$ are the reference block and its corresponding template area. Finally, the block with the smallest MSD is selected as the reference block of the current CU.
	
	\subsubsection{Edge detection}
	To reduce quantization noise and preserve object boundaries, we use the guided filter to replace the widely-used Gaussian filter in the canny detector, firstly. Then we conduct the remaining steps in the canny detector~(Sobel operator, non-maximum suppression and double threshold) to derive the edge map. The guided filter process is as follows,
	\begin{equation}
	q = a_kI + b_k,
	\end{equation}
	
	\begin{equation}
	a_k = \frac{\sigma_{k}^{2}}{\sigma_{k}^{2} + \epsilon},
	\end{equation}
	
	\begin{equation}
	b_k = (1 - a_k)u_k,
	\end{equation}
	
	\begin{equation}
	\epsilon = {\varepsilon \times (1 << bitDepth)}^2,
	\end{equation}
	
	\noindent where $I$ is the input reconstructed image, $q$ is the output of the guided filter. $\sigma_k$ is the variance of image $I$. $u_k$ is the mean of $I$. $bitDepth$ is the internal bit-depth, which is 10 in VVC. $\epsilon$ represents the filter strength, and $\varepsilon$~($\varepsilon \in [0, 1]$) is the parameter needed to be trained. Other than $\varepsilon$, two thresholds, $thre_{high}$ and $thre_{low}$, used in the double-threshold process of the canny detector also need to be trained offline. The actual high threshold~($THRE_{HIGH}$) used in the canny detector is the $k_{th}$ value in the list which is constructed by sorting the block amplitude map after non-maximum process according to the amplitude value from small to large, where $k$ is $w\times h \times thre_{high}$. The actual low threshold is calculated by $THRE_{HIGH}/thre_{low}$.
	
	\begin{table}[b!]
		\begin{center}
			\small
			\caption{MPM list construction.} \label{MPM}
			\begin{tabular}{m{1.6cm}<{\centering}|m{2.8cm}<{\centering}|m{2.8cm}<{\centering}}
				\thickhline
				\textbf{MPMIdx} & \textbf{AngleIdx}       & \textbf{DistIdx} \\
				\hline
				0      & bestAngIdx     & bestDistIdx \\
				\hline
				1      & bestAngIdx     & bestDistIdx - 1 \\
				\hline
				2      & bestAngIdx     & bestDistIdx + 1 \\
				\hline
				3      & bestAngIdx - 1 & bestDistIdx \\
				\hline
				4      & bestAngIdx - 1 & bestDistIdx - 1 \\
				\hline
				5      & bestAngIdx - 1 & bestDistIdx + 1 \\
				\hline
				6      & bestAngIdx + 1 & bestDistIdx \\
				\hline
				7      & bestAngIdx + 1 & bestDistIdx - 1 \\
				\hline
				8      & bestAngIdx + 1 & bestDistIdx + 1 \\
				\hline
				9      & default        & default \\
				\hline
				...    & ...            &...  \\
				\hline
				M      & default        & default \\
				\thickhline
			\end{tabular}
		\end{center}
		\vspace{-1mm}
	\end{table}
	
	\subsubsection{MPM list construction}
	After the reference block derivation and edge detection, a predicted edge map corresponding to the amplitude of Sobel gradient and an angle map corresponding to the direction of Sobel gradient are derived for the MPM list construction. The first step in MPM list construction process is to derive the best partition mode, corresponding to angle~(bestAngIdx) and distance~(bestDistIdx), with the largest average amplitude. Assuming the length of MPM list is $M$, the MPM list construction process is shown in Table.~\ref{MPM}. If the selected candidates do not fulfill the MPM list, a default value will be added in it. In this paper, we set the default value to be 0. 
	
	\begin{table}[ht!]
		\begin{center}
			\small
			\caption{Variance of edge detection parameters.} \label{train-var}
			\begin{tabular}{m{1.7cm}<{\centering}|m{1.7cm}<{\centering}|m{1.7cm}<{\centering}|m{1.7cm}<{\centering}}
				\thickhline
				& $\varepsilon$     & $thre_{high}$ & $thre_{low}$\\
				\hline
				PSNR-var      & 0.000529          & 0.000571      & 0.000359\\
				\hline
				bitRate-var   & 112.95            & 119.75        & 108.91\\
				\thickhline
			\end{tabular}
		\end{center}
		\vspace{-1mm}
	\end{table}

	\begin{figure}[hb!]
		\begin{center}
			\noindent
			\subfigure[$thre_{high}-QP$]{
				\includegraphics[width=1.5in]{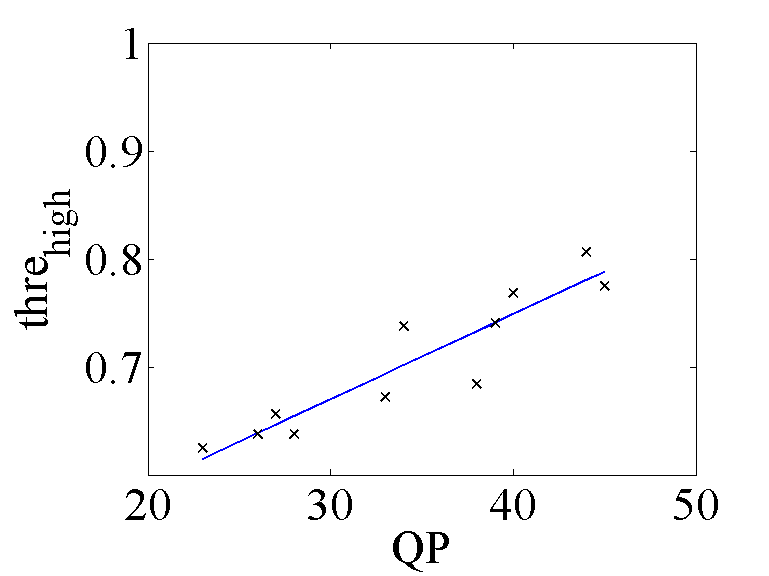}\label{thre}
			}
			\subfigure[$\varepsilon-QP$]{
				\includegraphics[width=1.5in]{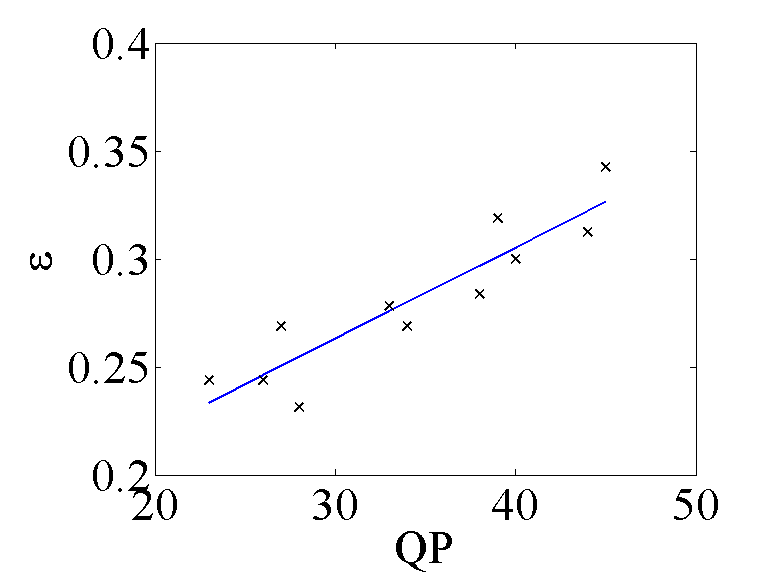}\label{eps_line}
			}
			\caption{Relationship between parameters and QP.}\label{training}
		\end{center}
		\vspace{-1mm}
	\end{figure}

	\subsubsection{Parameter training}
	The parameter $\varepsilon$ used in the guided filter, $thre_{high}$ and $thre_{low}$ applied in the canny detector are trained as follows,
	
	\begin{enumerate}	
		\item [1)] In this paper, we choose four non-test sequences~(\textit{BasketballPass}, \textit{BQSquare}, \textit{BlowingBubbles} and \textit{RaceHorses}), four typical QPs~(22, 27, 32, 37) and two configurations~(RA and LDB) for parameter training. The coding performance is tested with different parameter sets.
		
		\item [2)] According to the coding performance derived in previous step, we calculate the PSNR-variance and bitRate-variance by changing one parameter with the other two parameters fixed. As shown in Table.~\ref{train-var}, the parameter with larger variance represents that it has larger influence on coding efficiency. Based on this rule, we find that the most important parameter is $thre_{high}$, the second one is $\varepsilon$ and the third one is $thre_{low}$.
		
		\item [3)] For each B-frame, we experimentally select the optimal $thre_{high}$ corresponding to the highest PSNR of Y component, which is plotted in Fig.~\ref{thre}. Because of the QPOffset for different B-frame, there may be several QPs for one test configuration. Hence, the number of points in Fig.~\ref{training}~(a) is more than 4. It is clear that there is a positive relation between QP and the corresponding best $thre_{high}$. So we fit a linear model to represent their correlation. The fitted line, as shown in Fig.~\ref{thre}, can be formulated as,  
		%The most important parameter $thre_{high}$ is trained by selecting the optimal one for each B-frame and averaging these parameters to derive the optimal parameter for each QP. Because of the QPOffset for different B-frame, there may be several QPs for one test configuration. Hence, the number of points in Fig.~\ref{training}~(a) is more than 4. $thre_{high}$ can be formulated as,  
		\begin{equation}
		thre_{high} = 0.0079 \times QP + 0.4344\label{threHigh}.
		\end{equation}
		
		\item [4)] We train the parameter $\varepsilon$ as step 3) with $thre_{high}$ fixed using Eq.~\eqref{threHigh}. As shown in Fig.~\ref{eps_line}, the linear model for $\varepsilon-QP$ can be formulated as, 
		\begin{equation}
		\varepsilon = 0.0042 \times QP + 0.1363 \label{eps}.
		\end{equation}
		
		\item [5)] Finally, parameter $thre_{low}$ is trained with $thre_{high}$ and $\varepsilon$ fixed using Eq.~\eqref{threHigh} and Eq.~\eqref{eps}. As there is no obvious rule for this parameter, we set it to be 3.0 in this paper.
	\end{enumerate}

	\begin{table}[t!]
		\centering
		\small
		\setlength\tabcolsep{1mm}
		\begin{center}
			\caption{Experimental results of the proposed E-GEO, Anchor: VTM-6.0.} \label{Performance}
			\begin{tabular}{c|c|c|c|c|c}
				%\toprule[1.5pt]
				\thickhline
				\hline
				\multirow{2}{*}{\textbf{Class}}&
				\multirow{2}{*}{\textbf{Sequence}}&
				\multicolumn{2}{c|}{\rule{0pt}{8pt} \textbf{GEO}~\cite{P0068}} & \multicolumn{2}{c}{\textbf{Proposed E-GEO}}\\
				\cline{3-6} & & 
				\textbf{RA}& \textbf{LDB}& \textbf{RA}& \textbf{LDB}\\
				
				\hline
				\multirow{3}{*}{Class A1}
				& \rule{0pt}{8pt} \textit{Tango2}               & -0.23\% & -        & -0.30\%  & -  \\
				& \rule{0pt}{8pt} \textit{FoodMarket4}          & -0.11\% & -        & -0.16\%  & -  \\
				& \rule{0pt}{8pt} \textit{Campfire}             & -0.00\% & -        & -0.08\%  & -  \\
				
				\hline
				\multirow{3}{*}{Class A2}
				& \rule{0pt}{8pt} \textit{Catrobot}             & -0.31\% & -        & -0.41\%  & - \\
				& \rule{0pt}{8pt} \textit{DaylightRoad2}        & -0.19\% & -        & -0.32\%  & - \\
				& \rule{0pt}{8pt} \textit{ParkRunning3}         & -0.13\% & -        & -0.26\%  & - \\
				
				\hline
				\multirow{5}{*}{Class B}
				& \rule{0pt}{8pt} \textit{MarketPlace}          & -0.18\% & -0.51\%  & -0.37\%  & -0.74\% \\
				& \rule{0pt}{8pt} \textit{RitualDance}          & -0.53\% & -0.47\%  & -0.60\%  & -0.73\% \\
				& \rule{0pt}{8pt} \textit{Cactus}               & -0.25\% & -0.39\%  & -0.47\%  & -0.64\% \\
				& \rule{0pt}{8pt} \textit{BasketballDrive}      & -0.22\% & -0.30\%  & -0.36\%  & -0.46\% \\
				& \rule{0pt}{8pt} \textit{BQTerrace}            & -0.15\% & -0.02\%  & -0.24\%  & -0.24\% \\
				
				\hline
				\multirow{4}{*}{Class C}
				& \rule{0pt}{8pt} \textit{BasketballDrill}      & -0.46\% & -1.18\%  & -0.76\%  & -1.39\% \\
				& \rule{0pt}{8pt} \textit{BQMall}               & -1.79\% & -1.46\%  & -2.13\%  & -1.75\% \\
				& \rule{0pt}{8pt} \textit{Partyscene}           & -0.32\% & -0.46\%  & -0.52\%  & -0.70\% \\
				& \rule{0pt}{8pt} \textit{RaceHorsesC}          & -1.34\% & -0.92\%  & -1.70\%  & -1.34\% \\
				
				\hline
				\multirow{3}{*}{Class E}
				& \rule{0pt}{8pt} \textit{FourPeople}           & -       & -0.70\%  & -      & -1.04\% \\
				& \rule{0pt}{8pt} \textit{BQJohny}              & -       & -1.48\%  & -      & -1.86\% \\
				& \rule{0pt}{8pt} \textit{KristenAndSara}       & -       & -0.48\%  & -      & -1.07\% \\
				
				\hline
				\multirow{6}{*}{Average}
				& \rule{0pt}{8pt} {Class A1}           & -0.11\% & -        & -0.18\%      & -        \\
				& \rule{0pt}{8pt} {Class A2}           & -0.21\% & -        & -0.33\%      & -        \\
				& \rule{0pt}{8pt} {Class B}            & -0.27\% & -0.34\%  & -0.41\%      & -0.56\%  \\
				& \rule{0pt}{8pt} {Class C}            & -0.98\% & -1.00\%  & -1.28\%      & -1.30\%  \\
				& \rule{0pt}{8pt} {Class E}            &       - & -0.89\%  &  -           & -1.32\%  \\
				& \rule{0pt}{8pt} {Overall}            & -0.41\% & -0.70\%  &  -0.58\%     & -1.00\%  \\
				\thickhline
			\end{tabular}
		\end{center}
		\vspace{-1.8mm}
	\end{table}

	\begin{figure}[ht!]
		\begin{center}
			\noindent
			\subfigure[\textit{Anchor~(VTM-6.0)}]{
				\includegraphics[width=1.2in]{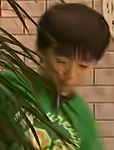}
			}
			\subfigure[\textit{Proposed E-GEO}]{
				\includegraphics[width=1.2in]{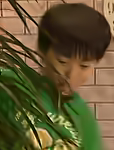}
			}
			\subfigure[\textit{Anchor~(VTM-6.0)}]{
				\includegraphics[width=1.2in]{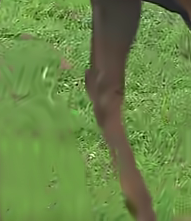}
			}
			\subfigure[\textit{Proposed E-GEO}]{
				\includegraphics[width=1.2in]{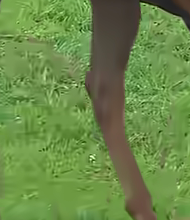}
			}
			\caption{Subjective results of the proposed E-GEO~(\textit{PartyScene} and \textit{RacehorsesC}, QP=37, LDB) }\label{subjective_result}
		\end{center}
		\vspace{-1.8mm}
	\end{figure}
	
	\begin{figure}[h]
		\begin{center}
			\noindent
			\includegraphics[width=2.4in]{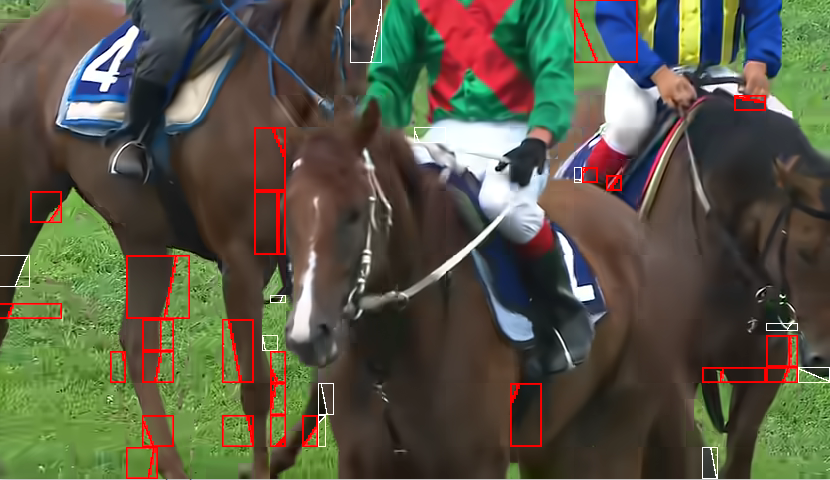}
			\caption{An example of the proposed E-GEO.~(\textit{RacehorsesC}, QP=37, the $58^{th}$ frame, RA)}\label{dec_partition_Pro}
		\end{center}
		\vspace{-1.8mm}
	\end{figure}
	
	\section{Experimental Results}
	In this section, we test the performance of the proposed E-GEO algorithm in VTM-6.0~\cite{VTM6.0}. The videos in the common test condition~\cite{CTC} are utilized as test sequences in our experiment under two configurations, Random Access Main10~(RA) and Low Delay B Main10~(LDB). The first two seconds of these sequences are encoded for performance evaluation. And the coding performance is measured by Bjontegaard's method~\cite{BDrate}~(Y-component) in terms of BD-rate.
	
	Table.~\ref{Performance} summarizes the results of the proposed method. Two tests are conducted, GEO~(CE4-1.1)~\cite{P0068} and the proposed E-GEO. It can be seen that the proposed method can achieve 0.58\% and 1.00\% bit-rate reduction for RA and LDB, which improves the coding gain of GEO by 41\% and 43\%. Note that the bit-rate reduction of class E in LDB configuration is higher. The reason is that the sequences of class E are conference videos with stable camera. In that case, the edge detection method is more accurate. E-GEO can also improve the subjective performance, as shown in Fig.~\ref{subjective_result}. Compared to VTM-6.0, E-GEO can improve subjective quality obviously, especially on the boundaries of moving objects. We also analyze the partition mode in the decoder, which is shown in Fig.~\ref{dec_partition_Pro}. In which, red squares represent that the selected GEO mode is in the MPM list and white squares represent the contrary. It can be seen that the proposed method can predict the GEO partition mode effectively. Moreover, there is still space for optimization. And other GEO mode prediction methods will be investigated in our future work.

	\section{Conclusion}
	In this paper, an MPM-based GEO scheme was proposed to reduce the overhead of GEO index and improve coding efficiency. Based on the observation of the high correlation between partition mode and object boundaries, an edge-directed geometric partition scheme was adopted to construct the MPM list according to the spatio-temporal predicted edge information. Experimental results demonstrated that our proposed algorithm can predict the GEO partition mode efficiently and achieve 0.58\% and 1.00\% bit-rate reduction on average for RA and LDB configurations, respectively.
	
	\section{ACKNOWLEDGEMENT}
	This work was supported by National Key Research and Development Project~(2019YFF0302703), Guangdong Key Research and Development Project~(2019B010133001), and High-performance Computing Platform of Peking University, which are gratefully acknowledged.
	
	% References should be produced using the bibtex program from suitable
	% BiBTeX files (here: strings, refs, manuals). The IEEEbib.bst bibliography
	% style file from IEEE produces unsorted bibliography list.
	% -------------------------------------------------------------------------
	\bibliographystyle{IEEEbib}
	\bibliography{icme2020template}
	
\end{document}